\documentclass[12pt]{article}

\usepackage{verbatim}

\usepackage{xr}
\usepackage[margin=0.9in]{geometry}
\usepackage{verbatim}
\usepackage{scicite}
\usepackage{todonotes}

\newcommand{\Armando}[1]{\todo[author=Armando]{#1}}

\usepackage{times}

\usepackage{multibib}

\usepackage{dblfloatfix}    % To enable figures at the bottom of page
\usepackage{microtype}
\usepackage{graphicx}
\usepackage{booktabs} \usepackage{stmaryrd}% for professional tables

\usepackage{verbatim}
\usepackage{longtable}
\usepackage[section]{placeins}
\newcommand{\dcaption}[1]{\caption{#1}}
\newcommand{\system}{DreamCoder~}
\newcommand{\systemEnding}{DreamCoder}

\newcommand{\code}[1]{{\footnotesize\texttt{#1}}}

\newcommand{\program}{\rho}
\newcommand{\prior}{L}
\newcommand{\library}{D}

\usepackage{xcolor}
\definecolor{pop1}{HTML}{1F78b4}
\definecolor{pop2}{HTML}{164C13}
\definecolor{pop3}{HTML}{d95F02}
\definecolor{orange}{HTML}{d95F02}
\definecolor{teal}{HTML}{1b9e77}

\usepackage{url}            % simple URL typesetting
\usepackage{booktabs}       % professional-quality tables
\usepackage{amsfonts}       % blackboard math symbols
\usepackage{nicefrac}       % compact symbols for 1/2, etc.
\usepackage{microtype}      % microtypography
\usepackage{mathrsfs}
\usepackage{listings}
\usepackage{amsthm}
\usepackage{times}
\usepackage{graphicx} % more modern
\usepackage{subfig} 
\usepackage{fancyvrb}

\usepackage{caption}

\fvset{fontsize=\footnotesize}

\usepackage{amssymb}
\usepackage{listings}
\usepackage{wrapfig}
\usepackage{tabularx}

 \usepackage{booktabs}
\usepackage{algorithm}
\usepackage{algpseudocode}% http://ctan.org/pkg/algorithmicx
\usepackage{tikz}
\usepackage{circuitikz}
\usetikzlibrary{fit,bayesnet,calc,tikzmark}
\usetikzlibrary{arrows.meta}
\usetikzlibrary{positioning}
\usetikzlibrary{decorations.text}

\usepackage{dsfont}
\usepackage{amsmath}

\DeclareMathOperator*{\argmax}{arg\,max} % thin space, limits underneath in displays

\newcommand{\expect}{\text{E}} %{{\rm I\kern-.3em E}}
\newcommand{\probability}{\text{P}} %{{\rm I\kern-.3em P}}

\newenvironment{sciabstract}{%
  \begin{quote} \bf}
{\end{quote}}

\title{%\vspace{-0cm}{\normalsize\color{red}{\emph{Draft in progress; subject to changes.}}}\\
  DreamCoder: Growing generalizable, interpretable knowledge with wake-sleep Bayesian program learning}

\author{Kevin Ellis,\textsuperscript{1,4,5} Catherine Wong,\textsuperscript{1,4,5} Maxwell Nye,\textsuperscript{1,4,5} \\ Mathias Sabl\'e-Meyer,\textsuperscript{1,3}  Luc Cary,\textsuperscript{1} Lucas Morales,\textsuperscript{1,4,6} Luke Hewitt,\textsuperscript{1,4,5}\\Armando Solar-Lezama,\textsuperscript{1,2,6} Joshua B. Tenenbaum\textsuperscript{1,2,4,5}\\%
\small  \textsuperscript{1}MIT\phantom{tt}\textsuperscript{2}CSAIL\phantom{tt}\textsuperscript{3}NeuroSpin%\\%
\small  \phantom{tt}\textsuperscript{4}Center for Brains, Minds, and Machines\\%
\small\textsuperscript{5}Department of Brain and Cognitive Sciences\phantom{tt}%\\%
\small\textsuperscript{6}Department of Electrical Engineering and Computer Science%\\%\phantom{tt}
}
\date{}
\begin{document}

\maketitle{}

\begin{sciabstract}
Expert problem-solving is driven by powerful languages for thinking about problems and their solutions%% , from physics to mathematics to design
. Acquiring expertise means learning these languages --- systems of concepts, alongside the skills to use them. We present DreamCoder, a system that learns to solve problems by writing programs. It builds expertise by creating programming languages for expressing domain concepts,
together with neural networks to guide the search for programs within these languages. A ``wake-sleep'' learning algorithm %interleaves two sleep learning phases, alternately extending %DreamCoder's 
alternately extends the language with new symbolic abstractions and %training 
trains the neural network on imagined and replayed problems. DreamCoder solves both classic inductive programming tasks and creative tasks such as drawing pictures and building scenes. It rediscovers the basics of modern functional programming, vector algebra and classical physics, including Newton's and Coulomb's laws. 
%DreamCoder grows its languages 
Concepts are built compositionally from %concepts
those learned earlier, yielding multi-layered symbolic representations that are interpretable and transferrable to new tasks, while still growing scalably and flexibly with experience.
%
%DreamCoder grows its languages compositionally from concepts learned earlier, building multi-layered symbolic representations interpretability and generalizability of symbolic approaches with the flexibility and scalability of neural approaches.

\end{sciabstract}

A longstanding dream in artificial intelligence (AI) has been to build a machine that learns like a child~\cite{turing50} -- that grows into all the knowledge a human adult does, starting from much less.
This dream remains far off, as human intelligence rests on many learning capacities not yet captured in artificial systems.
While machines %learning algorithms %today 
are typically designed for a single class of tasks, humans %can %
learn to solve an endless range and variety of problems, from cooking to calculus to graphic design. %Remarkably, %at least by the standards of contemporary machine learning, 
%Relative to contemporary machines, 
While machine learning is data hungry, typically generalizing weakly from experience, human learners can often generalize strongly from only modest % a modest amount of 
experience. % to learn any one new %individual 
%task.
Perhaps most distinctively, %human learners 
humans build expertise: 
%The knowledge they acquire %from experience 
We acquire knowledge that can be communicated and extended, 
growing new concepts on those built previously to become better and faster learners the more we master a domain.
%and as we learn more in a domain, they build new concepts hierarchically on those built previously, %thereby 
%becoming better at solving new problems they have not trained on. 
%
%These hallmarks of human learning --- broad applicability, sample efficiency, interpretable reuse of knowledge, and the growth of expertise --- have been targeted in different ways across different machine learning traditions
%
%In the quest to build more human-like learning in machines, different approaches have set their sights on one or another of these abilities 
%
% Different machine learning traditions have aspired to capture different subsets of these abilities 
%

This paper presents DreamCoder, a machine learning system that aims to take a step closer to these human abilities -- to efficiently discover interpretable, reusable, and generalizable knowledge across a broad range of domains.  DreamCoder embodies an approach we call ``wake-sleep Bayesian program induction'', and the rest of this introduction explains the key ideas underlying it: what it means to view learning as program induction, why it is valuable to cast program induction as inference in a %hierarchical
Bayesian model, and how a ``wake-sleep'' algorithm enables the model to grow with experience, learning to learn more efficiently in ways that make the approach practical and scalable. 

Our formulation of learning as program induction %(or inductive program synthesis)
traces back to the earliest days of AI~\cite{solomonoff1964formal}: We treat learning a new task as search for a program that solves it, or which has intended behavior.
%can generate the examples observed for that task.
Fig.~1 shows examples of program induction tasks in eight different domains that DreamCoder is applied to (Fig.~1A), along with an in-depth illustration of one task in the classic list-processing domain: learning a program that sorts lists of numbers (Fig.~1B), given a handful of input-output examples.
Relative to %% neural networks or other
purely statistical approaches, viewing learning as program induction brings certain advantages.
Symbolic programs exhibit strong generalization properties--intuitively, they tend to extrapolate rather than merely interpolate. This also makes learning very sample-efficient: Just a few examples are often sufficient to specify any one function to be learned.  By design, programs are richly human-interpretable: They subsume our standard modeling languages from science and engineering, and they expose knowledge that can be reused and composed to solve increasingly complex tasks.  Finally, programs are universal: in principle, any Turing-complete language %% -- DreamCoder adopts a version of the functional programming language Lisp --
can represent solutions to the full range of computational problems solvable by intelligence.  

Yet for all these strengths, and successful applications in a number of domains~\cite{liang11dcs, kulkarni2015picture, lake2015human, chater2013programs, gulwani2011automating,lazaro2019beyond,devlin2017neural}, program induction has had relatively limited impact in AI.
A Bayesian formulation helps to clarify the challenges, as well as a path to solving them. The programming language we search in specifies the hypothesis space and prior for learning; the shorter a program is in that language, the higher its prior probability.  While any general programming language can support program induction, previous systems have typically found it essential to start with a carefully engineered domain-specific language (DSL),
%a rich set of primitives that
which imparts a strong, hand-tuned inductive bias or prior.
Without a DSL the programs to be discovered would be prohibitively long (low prior probability), and too hard to discover in reasonable search times.
Even with a carefully tuned prior, though, search for the best program has almost always been intractable for general-purpose algorithms, because of the combinatorial nature of the search space.
%This is an instance of the more general computational intractability of Bayesian inference in combinatorial spaces (cite).
Hence most practical applications of program induction require not only a
hand-designed DSL but also a search algorithm hand-designed to exploit that DSL for fast inference. Both these requirements limit the scalability and broad applicability of program induction.

DreamCoder addresses both of these bottlenecks by learning %for itself the expertise needed
to compactly represent and efficiently induce programs in a given domain.
The system learns to learn -- to write better programs, and to search for them more efficiently -- by jointly growing two distinct kinds of domain expertise: (1) explicit declarative knowledge, in the form of a learned domain-specific language, capturing conceptual abstractions common across tasks in a domain; and (2) implicit procedural knowledge, in the form of a neural network that guides how to use the learned language to solve new tasks, embodied by a learned domain-specific search strategy. In Bayesian terms, the system learns both a prior on programs, and an inference algorithm (parameterized by a neural network) to efficiently approximate the posterior on programs conditioned on observed task data.

%\textbf{I wonder if we can get rid of this} This factorization is loosely inspired by dual-process models in cognitive science ~\cite{evans1984heuristic,kahneman2011thinking} and the study of human expertise~\cite{chi1981categorization,chi1988nature}. Human experts learn both declarative domain concepts that they can talk about in words -- artists learn arcs, symmetries, and perspectives; physicists learn inner products, vector fields, and inverse square laws -- as well procedural (and implicit) skill in deploying those concepts quickly to solve new problems.
%Together, these two kinds of knowledge let experts more faithfully classify problems based on the ``deep structure'' of their solutions~\cite{chi1981categorization,chi1988nature}, and intuit which concepts are likely to be useful in solving a task even before they start searching for a solution.

DreamCoder learns both these ingredients %% its explicit and implicit expertise -- both the domain-specific programming language and the approximate posterior --
in a self-supervised, bootstrapping fashion, growing them jointly across repeated encounters with a set of training tasks. % in a form of meta-learning.
This allows learning to scale to %arbitrary
new domains, and to scale within a domain provided it receives sufficiently varied training tasks. Typically only a moderate number of tasks suffices to bootstrap learning in a new domain. For example, the list sorting function in Fig.~1B represents one of 109 tasks that the system cycles through, learning as it goes to construct a library of around 20 basic operations for lists of numbers which in turn become components for solving many new tasks it will encounter.

DreamCoder's learned languages take the form of multilayered hierarchies of abstraction (Fig.~1B, \& Fig.~\ref{librariesToLanguages}A,B).  These hierarchies are reminiscent of the internal representations in a deep neural network, but here each layer is built from symbolic code defined in terms of earlier code layers, making the representations naturally interpretable and explainable by humans.  The network of abstractions grows progressively over time, building each concept on those acquired before, inspired by how humans build  conceptual systems: we learn algebra before calculus, and only after arithmetic; we learn to draw simple shapes before more complex designs. For example, in the list processing example (Fig.~1B), our model comes to sort sequences of numbers by invoking a library component four layers deep -- take the $n^{th}$ largest element -- and this component in turn calls lower-level learned concepts: maximum, and filter. Equivalent programs could in principle be written in the starting language, but those produced by the final learned language are more interpretable and much shorter. Expressed only in the initial primitives, these programs would be so complex as to be effectively out of the learner’s reach: they would never be found during a reasonably bounded search. Only with acquired domain-specific expertise do most problems become practically solvable.

\begin{figure}[t!]\centering
%  \includegraphics[width = \textwidth]{tasks.pdf}
  %  \includegraphics[width = \textwidth]{figures/logo_list_combined_task_bar_10_30.png}
  %% \includegraphics[width = \textwidth]{figures/main_figure_11_22-crop.pdf}\\\vfill

  %% \vspace{2cm}
  
% \includegraphics[width = \textwidth]{figures/main_figure_11_22-crop.pdf}
 \includegraphics[width = \textwidth]{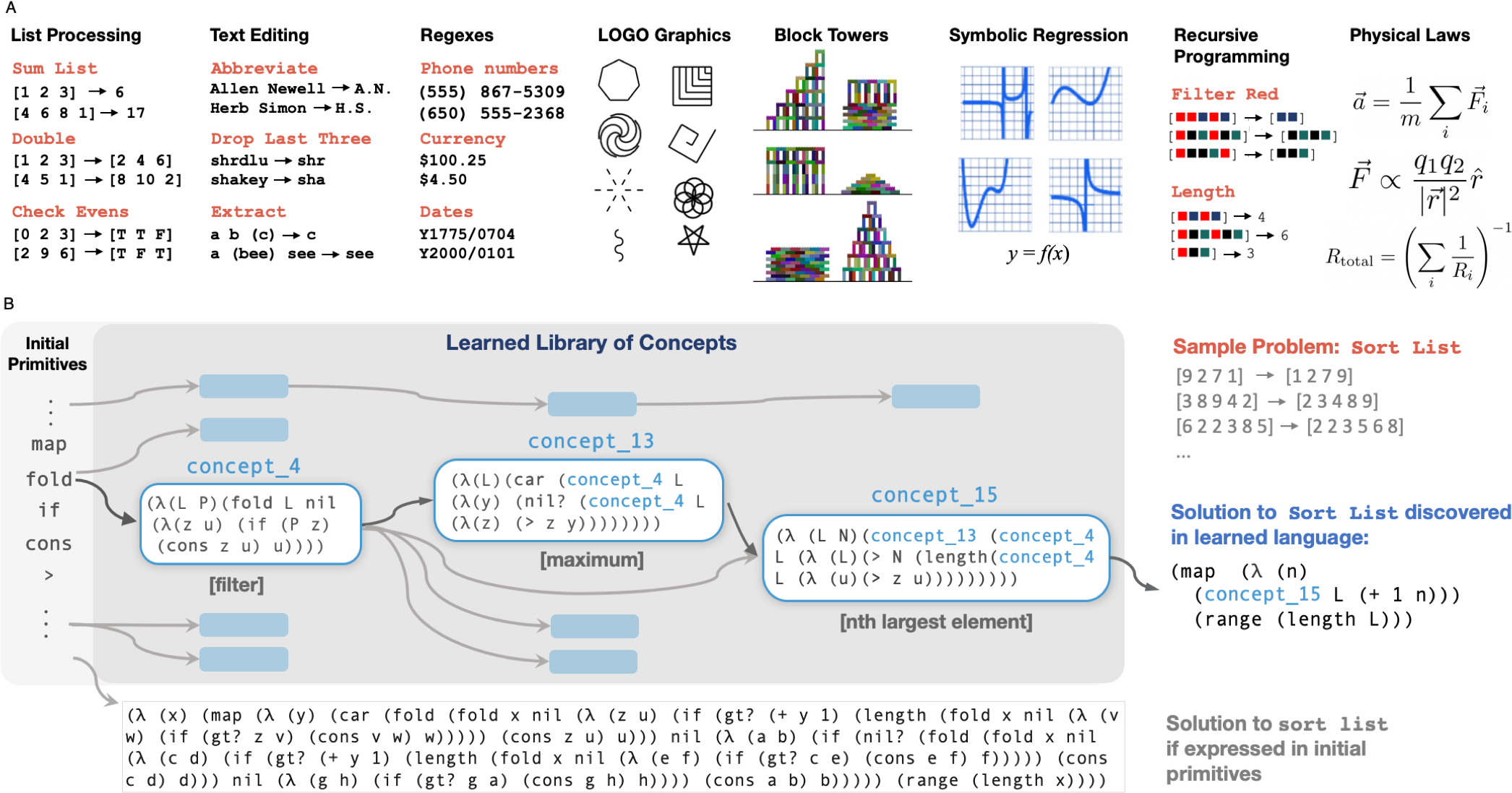}

 \dcaption{\textbf{(A)}: Learning tasks in many different domains can be formulated as inducing a program that explains a small number of input-output examples, or that generates an observed sequence, image or scene. DreamCoder successfully learns to synthesize programs for new tasks in each of these domains.  \textbf{(B)}: An illustration of how \system learns to solve problems in one domain, processing lists of integers. Problems are specified by input-output pairs exemplifying a target function (e.g., `Sort List').  Given initial primitives (left), the model iteratively builds a library of more advanced functions (middle) and uses this library to %write programs that 
 solve problems too complex to be solved initially.  Each learned function can call functions learned earlier (arrows), forming hierarchically organized layers of concepts. The learned library enables simpler, faster, and more interpretable problem solving: 
 % The learned library facilitates simpler and more interpretable solutions, and is a prerequisite for tractable problem solving: 
 A typical solution to `Sort List' (right), discovered after six iterations of learning, can be expressed with just five function calls using the learned library and is found in less than 10 minutes of search.  The code reads naturally as ``get the $n^{\text{th}}$ largest number, for $n = 1,2,3,\ldots $.''
   At bottom the model's solution is re-expressed in terms of only the initial primitives, yielding a long and cryptic program with 32 function calls, which would take in excess of $10^{72}$ years of brute-force 
   search to discover.   
  }\label{exampleDSL}
\end{figure}

DreamCoder gets its name from how it grows domain knowledge iteratively, in ``wake-sleep'' cycles loosely inspired by the memory consolidation processes that occur during different stages of sleep~\cite{DUDAI201520,fosse2003dreaming}. In general, wake-sleep Bayesian learning 	
~\cite{hinton1995wake} iterates between training a probabilistic \emph{generative model} that defines the learner's prior alongside a neural network \emph{recognition model} that learns to invert this generative model given new data. During ``waking'' the generative model is used to interpret new data, guided by the recognition model.  The recognition model is learned offline during ``sleep,'' from imagined data sets (``dreams'' or ``fantasies'') sampled %probabilistically
from the generative model. 

DreamCoder develops the wake-sleep approach for learning to learn programs: %that solve new tasks: 
Its learned language defines a generative model over programs and tasks, where each program solves a particular hypothetical task; its neural network learns to recognize patterns across tasks in order to best predict program components likely to solve any given new task.  
During waking, the system
is presented with data from several %new 
tasks and attempts to synthesize programs that solve then, using the neural recognition model to propose candidate programs.   %These candidates are then tested to see how well they do, in fact, solve the presented tasks. %These candidates are then scored according to 
%
% (in terms of their posterior probability under the generative model).
%The neural recognition model selects program components not only from primitives the system begins with, but from the full set of learned primitives embodied in the programming language that grows across iterations.
Learning occurs during two distinct but interleaved sleep phases,
% single sentence version:
alternately growing the learned language (generative model) by consolidating %out 
new abstractions 
%abstracting out 
from programs found %to have solved tasks 
during waking, and training the neural network (recognition model) on %replayed experiences from waking as well as  
``fantasy'' programs sampled from the generative model.
% multiple sentence version
%during \emph{abstraction sleep} the model grows its learned language (generative model) by replaying programs found during waking and refactoring them to expose new abstractions; while during \emph{dream sleep} the neural recognition model learns to use the learned language on new problems by training on replayed experiences from waking as well as ``dreams,'' or sampled programs.
%During what we call \emph{dream sleep} the neural recognition model is trained on 
%
%
%The neural network is trained on both actual tasks that have been solved, as well as samples, or ``dreams,'' from the learned language.  The neural network and %
This wake-sleep architecture builds on and further integrates a pair of ideas, Bayesian multitask program learning~\cite{Dechter:2013:BLV:2540128.2540316,DBLP:conf/icml/LiangJK10,lake2015human} and neurally-guided program synthesis~\cite{balog2016deepcoder,devlin2017robustfill}, which have been separately influential in the recent literature but have only been brought together in our work starting with the EC$^2$ algorithm~\cite{ecc}, and now made much more scalable in DreamCoder (see~\ref{relatedSupplement} for further discussion of prior work).

The resulting system has wide applicability. We describe applications to eight domains (Fig.~1A): classic program synthesis challenges, more creative visual drawing and building problems, and finally, library learning that captures the basic languages of recursive programming, vector algebra, and physics.
% illustrates eight domains where we have applied the model where we have applied our model, along with the learned program libraries that represent the expertise developed in two of these domains (Fig.~1B,C).  %These applications include classic program synthesis tasks of manipulating sequences of numbers and text, as well as more creative visual tasks of drawing pictures and building towers out of toy blocks, and even learning libraries that capture the basic languages of recursive programming, vector algebra and physics. %
All of our tasks involve inducing programs from very minimal data, e.g., five to ten examples of a new concept or function, or a single image or scene depicting a new object.  The learned 
languages span deterministic and probabilistic programs, and programs that act both generatively (e.g., producing an artifact like an image or plan) and conditionally (e.g., mapping inputs to outputs). Taken together, we hope these applications illustrate the potential for program induction to become a practical, general-purpose, and data-efficient approach to building intepretable, reusable knowledge in artificial intelligence systems.

\section*{Wake/Sleep Program Learning}\label{overviewSection}

\begin{figure}
  \includegraphics[width = \textwidth]{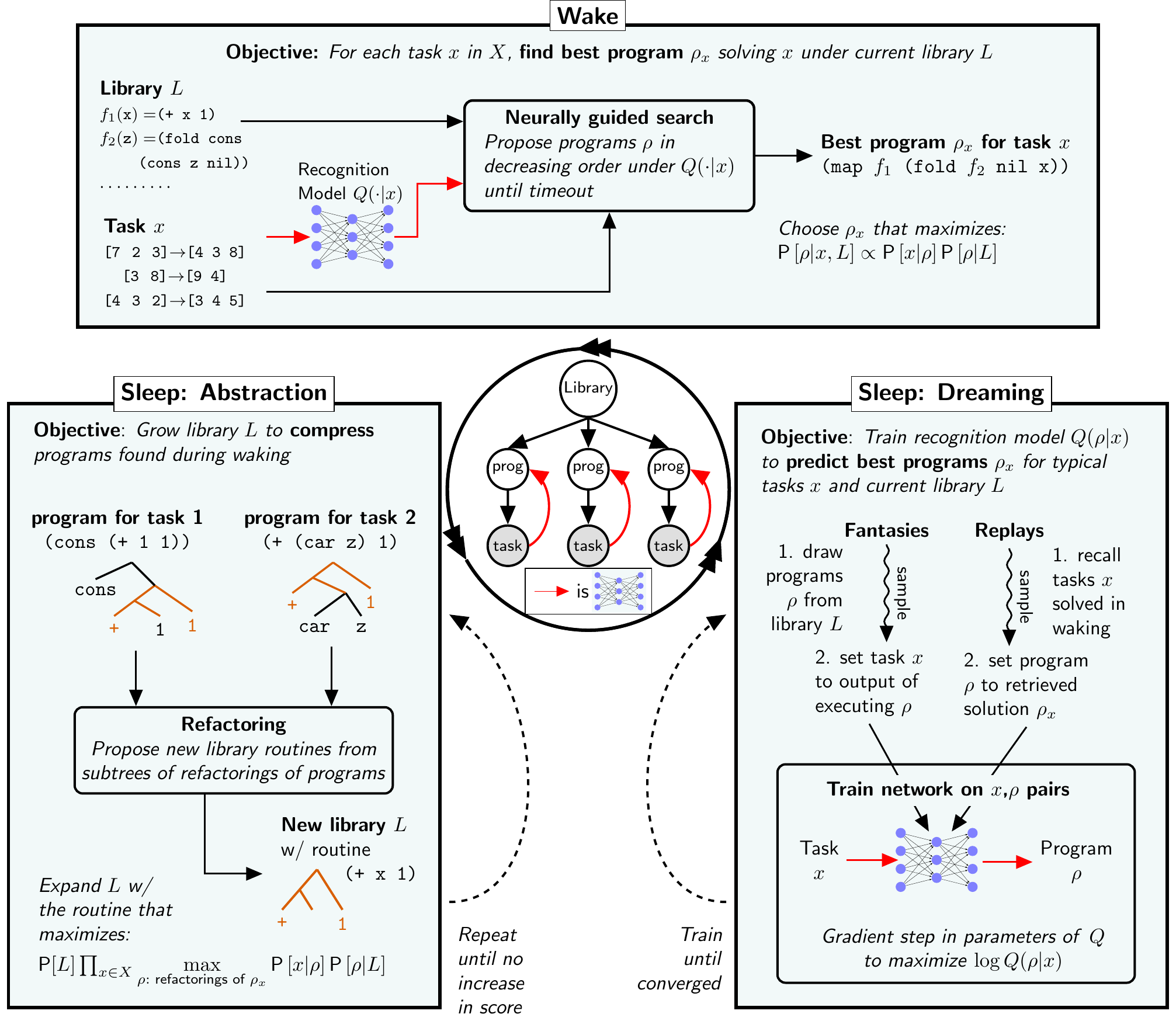}%cycleGraphicalModelCode}
  \caption{DreamCoder's basic algorithmic cycle, which serves to perform approximate Bayesian inference for the graphical model diagrammed in the \textbf{middle}. The system observes programming tasks (e.g., input/outputs for list processing or images for graphics programs), which it explains with latent programs, while jointly inferring a latent library capturing cross-program regularities. A neural network, called the \emph{recognition model} (red arrows) is trained to quickly infer programs with high posterior probability. The Wake phase (\textbf{top}) infers programs while holding the library and recognition model fixed. A single task, `increment and reverse list', is shown here. The Abstraction phase of sleep (\textbf{left}) updates the library while holding the programs fixed by refactoring programs found during waking and abstracting out common components (highlighted in orange). Program components that best increase a Bayesian objective (intuitively, that best compress programs found during waking) are incorporated into the library, until no further increase in probability is possible. A second sleep phase, Dreaming (\textbf{right}) trains the recognition model to predict an approximate posterior over programs conditioned on a task. The recognition network is trained on `Fantasies' (programs sampled from library) and `Replays' (programs found during waking).}\label{threeCycles}
\end{figure}

We now describe the specifics of learning in DreamCoder, beginning with an overview of the algorithm and its mathematical formulation, then turning to the details of its three phases.  Learning proceeds iteratively, %through a novel kind of ``wake-sleep'' learning, inspired by but crucially different than the original wake-sleep algorithm of Hinton, Dayan and colleagues~\cite{hinton1995wake}.
%% \begin{comment}
%%   For us, ``waking'' corresponds to solving problems by writing programs, while a pair of ``sleep'' cycles correspond to improving the library of explicit concepts -- increasing the breadth and depth of networks like those in Fig.~\ref{exampleDSL} -- and training the neural net to improve implicit procedural skill, respectively.
%% \end{comment}
%
%For explicit declarative knowledge, the challenge is to find a richer representational basis such that solutions to problems in the domain are easy to express. But enriching this representation poses a problem for search, because it grows the space of problems we can effectively solve with a small amount of code. Thus the procedural skill grows even more important as we build a richer base of declarative expert knowledge.
with each iteration (Eq.~\ref{threeEquations}, Fig.~\ref{threeCycles}) cycling through a wake phase of trying to solve tasks %--- where the model solves tasks by writing programs --- %wake cycle of solving a random sample of problems presented to the learner, %where it solves  tasks by writing programs (Fig.~\ref{threeCycles} top),
interleaved with two sleep phases for learning to solve new tasks. %; the pair of sleep stages alternatingly build declarative and procedural expertise.
%building explicit declarative knowledge and implicit procedural skills, respectively. % 
In the {\bf wake} phase (Fig.~\ref{threeCycles} top), the system searches for programs that solve tasks drawn from a training set, %either the full set of training tasks or a random subset (``minibatch'') of tasks, depending on the domain's size and complexity.  
guided by the neural recognition model which ranks candidate programs based on the observed data for each task. %These candidates are then tested to see how well they do, in fact, solve the presented tasks. 
Candidate programs are scored according to how well they solve the presented tasks, and how plausible they are a priori under the learned generative model for programs.  
%
% (in terms of their posterior probability under the generative model).
The first sleep phase, which we refer to as
\textbf{abstraction} (Fig.~\ref{threeCycles} left),
%inspired by abstraction processes during slow-wave sleep~\cite{DUDAI201520},
grows the library of programming primitives (the generative model) by replaying
experiences from waking, finding common program fragments from task solutions, and abstracting out these fragments into new code
primitives.
This mechanism increases the breadth and depth of the learner's declarative knowledge, its learned library as in Fig.~\ref{exampleDSL}B or Fig.~\ref{librariesToLanguages}, when viewed as a network.
%This cycle is loosely inspired by the formation of declarative abstractions during slow-wave sleep memory consolidation~\cite{DUDAI201520}.
The second sleep phase, which we refer to as \textbf{dreaming} (Fig.~\ref{threeCycles} right),
%inspired by replaying hallucination processes in REM sleep~\cite{fosse2003dreaming},
improves the
learner's procedural skill in code-writing by training the neural network that helps %quickly 
search for programs.
The neural recognition model is trained on replayed experiences as well as ``fantasies'', or programs sampled randomly from the learned library as a generative model. These random programs define tasks which the system solves during the dream phase, and the neural network is trained to predict the solutions found given the observable data for each imagined task.  
%Analogous processes of imagination and fast `pre-play' have recently been shown to occur during waking states as well~\cite{liu2019human,schuck2019sequential}.
%These two kinds of dreams are inspired by the distinct episodic replay and hallucination components of dream sleep~\cite{fosse2003dreaming}.

Viewed as a probabilistic inference problem, \system observes a training set of
tasks, written $X$, and infers both a program $\program_x$ solving each task $x\in X$, as
well as a prior distribution over programs likely to solve tasks in
the domain (Fig.~\ref{threeCycles} middle). This prior is encoded by
a library, written $\prior$, which %% combined with a
%% learned weight vector $\theta$
defines a generative model over programs,
written $\probability[\program|\prior]$
(see~\ref{generativeAppendix}).
%We can also treat $\prior$ as a prior over tasks by sampling a program from $\probability[\program|\prior]$ and then executing it to get a task.\footnote{Conditional programs--which map inputs to outputs--require also conditioning on example inputs, and we identify here $x$ with the set of inputs/outputs.}
The neural network %%  inverts
%% this generative model --- i.e.,
 helps to find programs solving a task by predicting, conditioned on the observed examples for that task, an approximate
posterior distribution over programs likely to solve it. %that specific
%task.
The network
thus functions as a \textbf{recognition model} that is trained jointly with the generative model, in the spirit of the Helmholtz machine~\cite{hinton1995wake}.
%The recognition model ensures that the combinatorial search over programs remains tractable, even as the library expands.
We write $Q(\program|x)$ for the approximate
posterior predicted by the recognition model.
At a high level wake/sleep cycles correspond to iterating the following updates, %% (Fig.~\ref{threeUpdates})
 illustrated in Fig.~\ref{threeCycles}; these updates serve to maximize a lower bound  on the posterior over $\prior$ given $X$ (\ref{probabilisticAppendix}).
\begin{align}
  \program_x &= \argmax_{\substack{\program:\\Q(\program|x)\text{ is large}}} \probability[\program|x,\prior] \propto \probability[x|\program]\probability[\program|\prior]\text{, for each task $x\in X$}&&\text{\emph{Wake}}\nonumber\\%\label{approximateWakingEquation}
  \prior &= \argmax_{\prior}\probability\left[\prior \right]\prod_{x\in X}\max_{\phantom{t}\program\text{ a refactoring of }\program_x\phantom{t}}\probability[x|\program]\probability[\program|\prior]&&\text{\emph{Sleep: Abstraction}}\nonumber\\%\label{sleepAbstractionObjective}
\text{Train }  Q(\program|x)&\approx \probability[\program|x,\prior]\text{, where $x\sim X$ (`replay') or $x\sim \prior$ (`fantasy')}&&\text{\emph{Sleep: Dreaming}} \label{threeEquations}%\label{approximateDreamingObjective} 
\end{align}
where $\probability[L]$ is a description-length prior over libraries (\ref{appendixCompression}) and  $\probability[x|\program]$ is the likelihood of a task $x\in X$ given program $\program$.
For example, this likelihood is 0 or 1 when $x$ is specified by inputs/outputs,
and when learning a probabilistic program, the likelihood is the probability of the program generating the observed task.
%% \begin{figure}[H]
%% \fbox{\includegraphics[width = \textwidth]{algorithmBox-crop.pdf}}
%%   \caption{DreamCoder's mathematical framing and parallel pseudocode. The model cycles through three updates (wake/abstraction/dreaming). Each update serves to maximize an \emph{objective} via an \emph{algorithm}, and together repeatedly solving for these objectives increases a lower bound on the posterior $\probability\left[L|X \right]$. Fig.~\ref{threeCycles} visually illustrates these cycles.}\label{threeUpdates}
%%   \end{figure}
%These updates serve to maximize a lower bound

This 3-phase inference procedure works through two distinct kinds of
bootstrapping.  During each sleep cycle the next library bootstraps
off the concepts learned during earlier cycles, growing an
increasingly deep learned library.  Simultaneously the
generative and recognition models bootstrap each other: A more specialized
library yields richer dreams for the recognition
model to learn from, while a more accurate recognition model solves
more tasks during waking which then feed into the next library.
Both sleep phases also serve to mitigate the combinatorial explosion
accompanying program synthesis.
Higher-level library routines allow tasks to be solved with fewer function calls,
effectively reducing the \emph{depth} of search.
The neural recognition model down-weights unlikely trajectories through the search space of all programs,
effectively reducing the \emph{breadth} of search.\footnote{We thank Sam Tenka for this observation. In particular, the difficulty of search during waking is roughly proportional to $\text{breadth}^\text{depth}$, where depth is the total size of a program and breadth is the number of library functions with high probability at each decision point in the search tree spanning the space of all programs. Library learning decreases depth at the expense of breadth, while training a neural recognition model effectively decreases breadth by decreasing the number of bits of entropy consumed by each decision (function call) made when constructing a program solving a task.}

{\bf Wake phase.} Waking consists of searching for task-specific programs with high
posterior probability, or programs that combine high likelihood (because they solve a task) and high prior probability (because they have short description length in the current language). 
During each Wake cycle we sample tasks from a random minibatch of the training set (or, depending on domain size and complexity, the entire training set). We then search for programs solving each of these tasks by enumerating programs in
decreasing order of their probability under the recognition model $Q(\program|x)$,
and checking if a program $\program$ assigns positive probability to solving that task
($\probability[x|\program] > 0$).
Because the model may find many programs that solve a specific task, % may have high posterior probability for a specific task,
we store a small beam of the $k=5$ programs with the highest posterior probability $\probability[\program|x,L]$, % per task,
and marginalize over this beam in the  sleep updates of Eq.~\ref{threeEquations}. %%. % of Eq.~\ref{threeEquations}.
We represent programs as polymorphically
typed $\lambda$-calculus expressions, an expressive formalism
including conditionals, variables, higher-order functions,
and the ability to define new functions.
%See supplement~\ref{enumerationAppendix} for details.

%\subsection{Abstraction Sleep: Growing a Deep Library of Concepts}\label{consolidationSection}

{\bf Abstraction phase.} During the abstraction sleep phase, %of sleep, 
the model grows its library of concepts with the goal of discovering specialized abstractions that allow it to easily
express solutions to the tasks at hand.
Ease of expression translates into a preference for libraries that best compress
programs found during waking,
and the abstraction sleep objective (Eq.~\ref{threeEquations})
is equivalent to minimizing the description  length of the library
($-\log \probability[\library]$)
plus the description lengths of
refactorings of programs found during waking
($\sum_x\min_{\program\text{ refactors }\program_x}-\log \probability[x|\program]\probability[\program|\library]$).
Intuitively,
we ``compress out'' reused code to maximize a Bayesian criterion,
but rather than compress out reused syntactic structures,
we refactor the programs to expose
reused semantic patterns.

\begin{figure}[t]
  \centering\includegraphics[width = 0.8\textwidth]{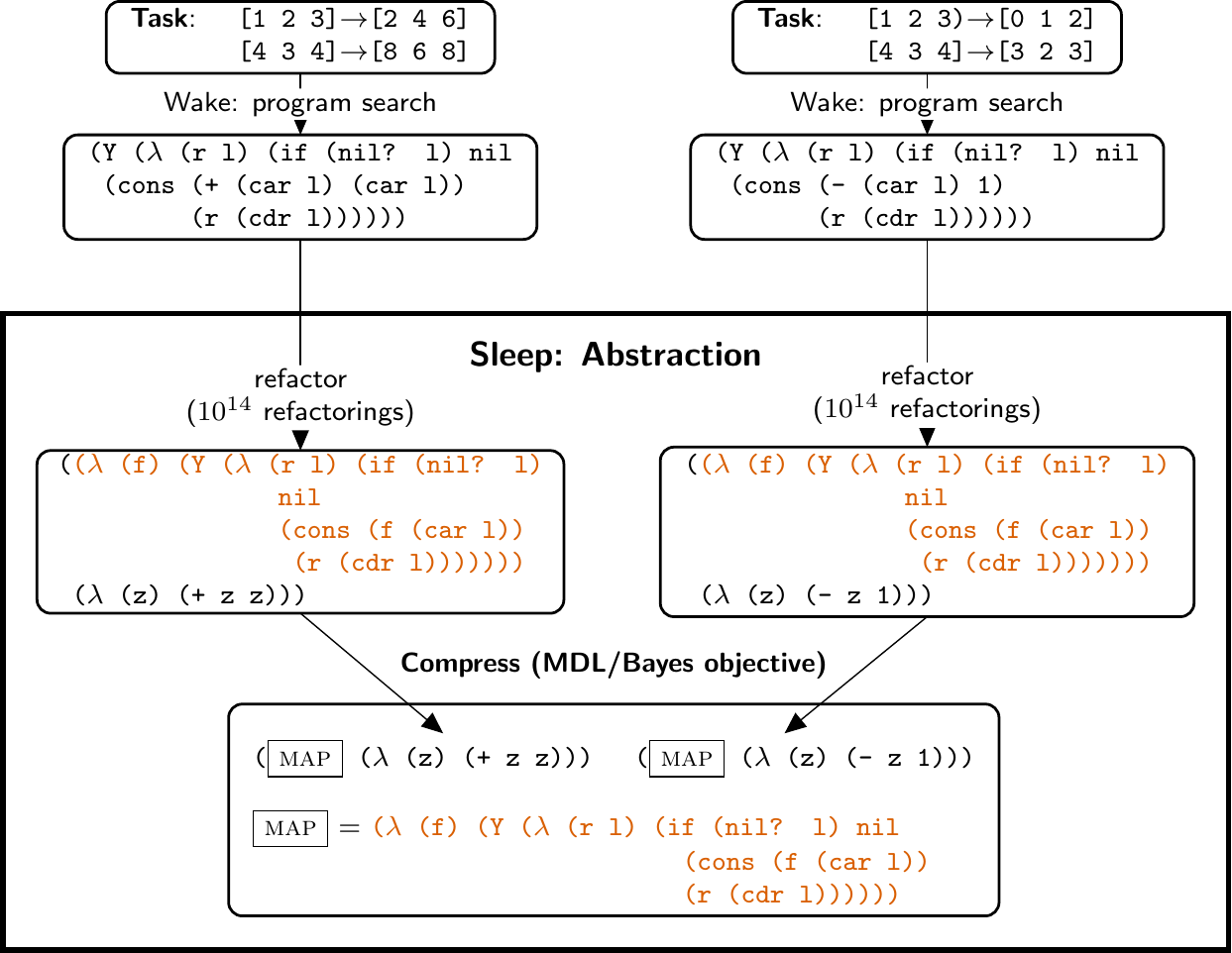}  \dcaption{
%  Library learning during the abstraction phase of sleep is driven by an efficient algorithm for refactoring programs found during waking. 
%In the sleep-abstraction phase of  learning, 
Programs found as solutions during waking are refactored -- or rewritten in semantically equivalent but syntactically distinct forms -- during the sleep abstraction phase, to expose candidate new primitives for growing DreamCoder's learned library. Here, solutions for two simple list tasks (top left, `double each list element'; top right, `subtract one from each list element') are first found using a very basic primitive set, which yields correct but inelegant programs. During sleep, DreamCoder efficiently searches an exponentially large space of refactorings for each program; a single refactoring of each is shown, with a common subexpression highlighted in orange. This %reused 
expression corresponds to \code{map}, a core higher-order function in modern functional programming that applies another function to each element of a list. Adding \code{map} to the library makes existing problem solutions shorter and more interpretable, and crucially bootstraps solutions to many harder problems in later wake cycles.}\label{mapFactor}
    %% Library learning as code refactoring. During waking the model solves tasks, such as `double each list element' (top left) and `subtract one from each list element' (top right),
    %% by synthesizing programs.
    %% Abstraction sleep refactors these programs to expose reused subexpressions (highlighted in %\orange{orange}
    %% orange) which are then added to the library, leading to shorter and more interpretable solutions (bottom).
    %% Here the discovered reused expression is `map,' which applies a function to each list element.
    
    %% Common subprograms are incorporated into the library when they increase a Bayesian objective. Intuitively, these new components best compress the programs found during waking.
\end{figure}

Code can be refactored in infinitely many ways,
so we bound the number of $\lambda$-calculus evaluation steps separating a program from its refactoring,
giving a finite but typically astronomically large set of refactorings.
Fig.~\ref{mapFactor} diagrams the model
discovering one of the most elemental building blocks of modern functional programming, the higher-order function
\code{map}, starting from a small set of universal primitives, including recursion (via the Y-combinator).
In this example there are approximately $10^{14}$ possible refactorings -- a quantity
that grows exponentially both as a function of program size and as a
function of the bound on evaluation steps.
To resolve this exponential growth we introduce a new data structure for representing and manipulating the set of refactorings, combining ideas from version space
algebras~\cite{lau2003programming,mitchell1977version,polozov2015flashmeta}
and equivalence graphs~\cite{tate2009equality}, %and give a dynamic program for constructing such a data structure in Appendix~\ref{appendixCompression}.
and we derive a dynamic program for its construction (supplementary~\ref{appendixCompression}).
This data structure grows polynomially with program size,
owing to a factored representation of shared subtrees,
but grows exponentially with a bound on evaluation steps,
and the exponential term can be made small (we set the bound to 3) without performance loss.
This results in substantial efficiency gains: A version space with $10^6$ nodes, calculated in minutes, can represent the $10^{14}$ refactorings in Fig.~\ref{mapFactor} that would otherwise take centuries to explicitly enumerate and search. 

% last kevin version
%\dcaption{Given programs found during waking (top left, solution for `double each list element'; top right, `subtract one from each list element'), the abstraction phase of sleep exhaustively yet efficiently searches an exponentially large space of program refactorings using a new refactoring algorithm. A single refactoring of each program is shown, with a common subexpression highlighted in orange. This reused subexpression is `map,' which applies a function to each element of a list. After adding `map' to the library the system's problem solutions are shorter and more interpretable. Discovering `map' also bootstraps the solving of harder problems in later wake cycles.

%\subsection{Dream Sleep: Training a Neural Recognition Model}\label{recognitionSection}

{\bf Dreaming phase.} During the dreaming sleep phase, the system trains its recognition model, which later speeds up problem-solving during waking
by guiding program search.
We implement recognition models as neural networks,
%which allows us to
injecting domain knowledge through the network architecture:
for instance, when inducing graphics programs from images, we use a convolutional network, which imparts a bias toward useful image features.
We train a recognition network on %is trained on
(program, task) pairs drawn from two
sources of self-supervised data: \emph{replays} of programs discovered
during waking, and \emph{fantasies}, or programs drawn from $\prior$.
Replays ensure that the recognition model is trained on the actual
tasks it needs to solve, and does not forget how to solve them,
while fantasies provide a large and highly
varied dataset to learn from,
and are critical for data efficiency:
becoming a domain expert is not a few-shot learning problem,
but neither is it a big data problem.
We typically train \system on 100-200 tasks,
which is too few examples for a high-capacity neural network.
After the model learns a library customized to the domain,
we can draw unlimited samples or `dreams' to train the recognition network.

Our dream phase works differently from a conventional wake-sleep~\cite{hinton1995wake} dream phase. %We think of dreaming as creating an endless stream of random problems, which we then solve during sleep, and train the recognition network to predict the solution conditioned on the problem. The classic wake-sleep algorithm would instead sample a random program, execute it to get a task, and train the recognition network to predict the sampled program from the sampled task.
A classic wake-sleep approach would sample a random program from the generative model,
execute it %on random inputs to get the outputs of a task,
to generate a task, and train the recognition network to predict the sampled program from the sampled task.
We instead think of dreaming as creating an endless stream of random problems,
which we then solve during sleep in an active process using the same program search process as in waking. We then train the recognition network to predict the solutions discovered, conditioned on the problems.
Specifically, we train $Q$ to perform MAP inference
by maximizing $\expect\left[\log Q\left(\left(\argmax_{\program}\probability[\program |x,\prior] \right)|x \right) \right]$,
where the expectation is taken over tasks. Taking this expectation over the empirical distribution of tasks trains $Q$ on replays; taking it over samples from the generative model trains $Q$ on fantasies. % (cf. Dyna:~\cite{sutton1991dyna}).
We train on a 50/50 mix of replays and fantasies;
for fantasies mapping inputs to outputs, we sample inputs from the training tasks.
Although one could
train $Q$ to perform full posterior inference, % by minimizing $\text{KL}\left(\probability[p|x,\prior]\|Q(p|x) \right)$,
%like in the classic wake-sleep algorithm,
our MAP objective has the advantage
of teaching the recognition network to find a simplest canonical solution for each problem.
More technically, our MAP objective acts to break syntactic symmetries
in the space of programs by forcing the network to place all its probability mass onto a single member of a set of
syntactically distinct but semantically equivalent expressions.
Hand-coded symmetry breaking has proved vital for many program synthesizers~\cite{solar2008program,feser2015synthesizing};
see~\ref{recognitionAppendix} for theoretical and empirical
analyses of \systemEnding's learned symmetry breaking.

\section*{Results}
We first experimentally investigate \system within two classic benchmark domains: list
processing and text editing. In both cases we solve tasks specified by
a conditional mapping (i.e., input/output examples), starting with a
generic functional programming basis, including routines like
\code{map}, \code{fold}, \code{cons}, \code{car}, \code{cdr}, etc.
% ec2 solves 94%
% 231 Lucas tasks - 21 identity = 210
% 105 training
% 105 test - 31 trivial = 74 test
% ec2 solved .94*105 = 98.7 solved, which would be 67.69 solved, or 91.49% accuracy on the difficult set of testing
% dc solves a median of 94%
%% : \code{foldr}, \code{unfold}, \code{if}, \code{map},
%% \code{length}, \code{index}, \code{=}, \code{+}, \code{-}, \code{0},
%% \code{1}, \code{cons}, \code{car}, \code{cdr}, \code{nil}, and
%% \code{is-nil}.
%\subsubsection{List Processing}\label{listSection}
Our list processing tasks comprise 218 problems taken from~\cite{ecc}, split 50/50 test/train,
each with 15 input/output examples.
In solving these problems, DreamCoder composed around 20 new library routines (\ref{deepMontageAppendix}), and rediscovered
higher-order functions such as \code{filter}. Each round of abstraction built on
concepts discovered in earlier sleep cycles --- for example the
model first learns \code{filter}, then uses
it to learn to take the maximum element of a list, then
uses that routine to learn a new library routine for extracting the
$n^{\text{th}}$ largest element of a list, which it finally uses to
sort lists of numbers (Fig.~1B).
%% This hierarchical, modular learning of concepts occurs because of the alternation between code writing
%% (during waking) and code refactoring (during the abstraction phase
%% of sleep).

%\subsubsection{Text Editing}\label{textSection}
Synthesizing programs that edit text is a classic problem in the
programming languages and AI literatures~\cite{lau2003programming},
and algorithms that synthesize text editing programs ship in Microsoft
Excel~\cite{gulwani2011automating}.
These systems would, for example, %These programs are synthesized from example input-output mappings,
%e.g.,
see the mapping $\text{``Alan Turing''}\to\text{``A.T.''}$,
and then infer a program that transforms ``Grace Hopper'' to ``G.H.''.
Prior text-editing program synthesizers rely on
hand-engineered libraries of primitives and hand-engineered search strategies.  Here, we
jointly learn both these ingredients and perform comparably to
a state-of-the-art domain-general program synthesizer. %%  on a standard
%% text editing benchmark.
We trained our system on 128 automatically-generated text editing
tasks, and tested %with 4 input/output examples each.  We tested, but did not
%train,
on the 108 text editing problems from the 2017
SyGuS~\cite{alur2017sygus} program synthesis
competition.\footnote{We compare with the 2017 benchmarks because 2018 onward introduced non-string manipulation problems; custom string solvers such as FlashFill~\cite{gulwani2011automating} and the latest custom SyGuS solvers are at ceiling for these newest problems.} Prior to learning, \system solves 3.7\% of
the problems within 10 minutes with an average search time of 235
seconds.  After learning, it solves 79.6\%, and does so much faster,
solving them in an average of 40 seconds.
%DreamCoder also converges $6\times$ faster than EC$^2$ yet solves more problems (\ref{appendixEC}).
The best-performing
synthesizer in this competition (CVC4) solved 82.4\% of the problems
--- but here, the competition conditions are 1 hour \& 8 CPUs per
problem, and with this more generous compute budget we solve 84.3\% of
the problems.  SyGuS additionally comes with a different
hand-engineered library of primitives \emph{for each text editing
  problem}. %% \footnote{SyGuS text editing problems also prespecify the
%% set of allowed string constants for each task. For these experiments,
%% our system did not use this assistance.}
Here we learned a single
  library of text-editing concepts that applied generically to any editing task, a prerequisite for real-world use.
%% \begin{figure}[h]\centering
%%   \begin{tabular}{ccc}
%%     \fbox{\begin{tabular}{c}
%%         \;\,Dream Coder$\to$(Coder)\\
%%         Domain Expertise$\to$(Expertise)
%%     \end{tabular}}&
%%     \fbox{\begin{tabular}{c}
%%         Temple Anna H$\to $Dr. TAH\\
%%         Lara Gregori$\to $Dr. LG
%%     \end{tabular}}&
%%     \fbox{\begin{tabular}{c}
%%       this test-split here.$\to$split here\\
%%       another-text.test for you$\to$text
%%     \end{tabular}}
%%   \end{tabular}
%%   \caption{Three example text editing tasks. Each task is specified by input/output pairs (notated above as input$\to$output), and the goal is to infer a program producing the input/output mapping.}\label{textEditExamples}
%% \end{figure}

%\section*{Generative \& Procedural Programs for making images and plans}
%% We consider three domains where the agent must infer a program from an
%% image (Fig.~\ref{visualSpecs}).
%% First we consider programs that make plans and take actions:  drawing pictures and building towers out of blocks (Sec.~\ref{logoSection}-\ref{towerSection}).

%\subsubsection{Programs that generate images}\label{logoSection}

\begin{figure}
  \centering
\vspace{-1cm}  \includegraphics[width = \textwidth]{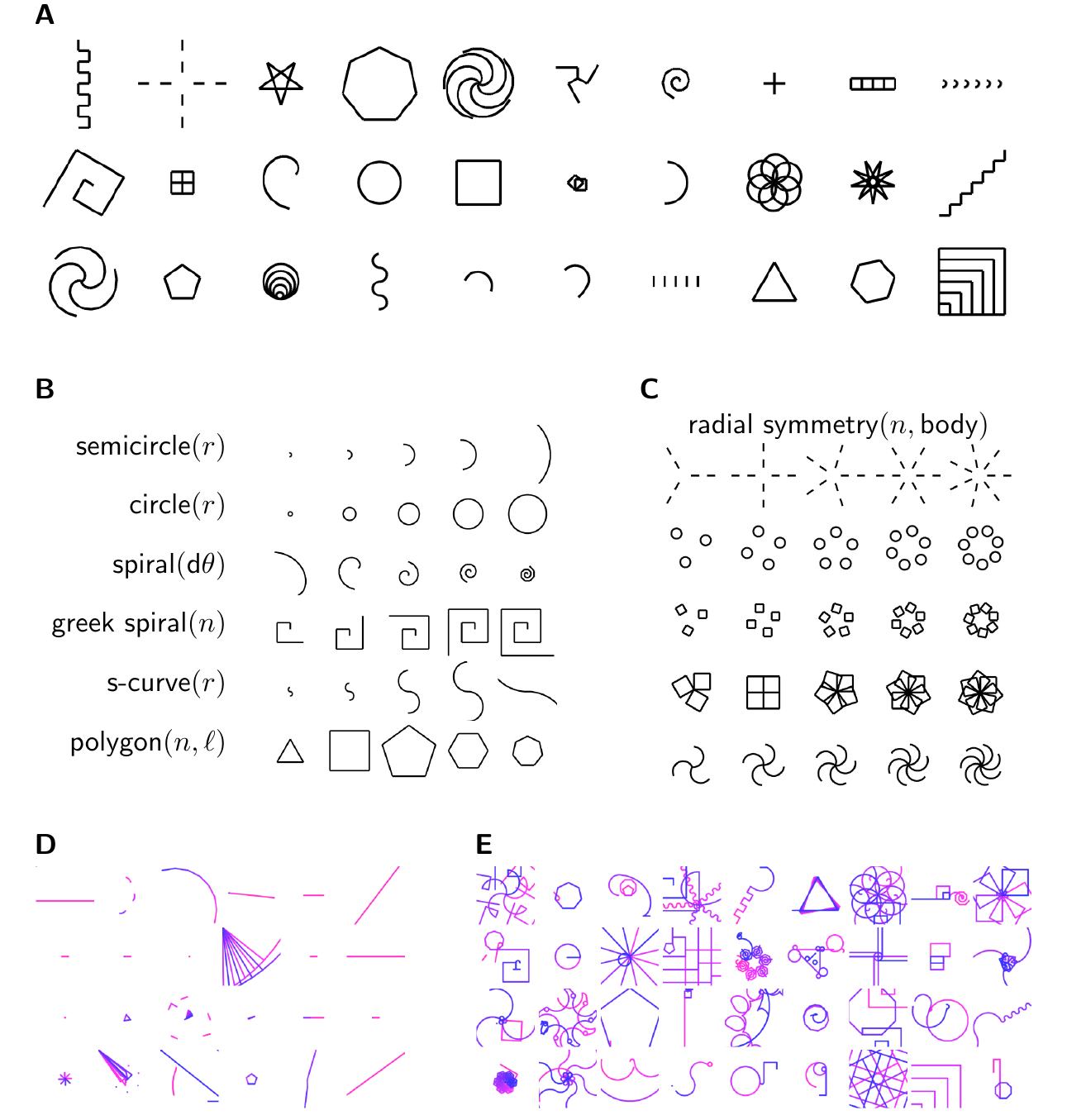}
  \dcaption{\textbf{(A):} 30 (out of 160) LOGO graphics tasks. The model writes programs controlling a `pen' that draws the target picture. \textbf{(B-C):} Example learned library routines %when trained on these images. The model learns 
  include both %interpretable 
  parametric routines for drawing families of curves (\textbf{B}) as well as primitives that take entire programs as input (\textbf{C}). Each row in \textbf{B} shows the same code executed with different parameters. Each image in \textbf{C} shows the same code executed with different parameters and a different subprogram as input.  \textbf{(D-E)}: Dreams, or programs sampled by randomly assembling functions from the model's library, change dramatically over the course of learning reflecting learned expertise. Before learning (\textbf{D}) dreams can use only a few simple drawing routines and are largely unstructured; the majority are simple line segments. After twenty iterations of wake-sleep learning (\textbf{E}) dreams become more complex by recombining learned library concepts in %new 
  ways never seen in the training tasks. Dreams are sampled from the  prior learned over tasks solved during waking, and provide an infinite stream of data for training the neural recognition model. % to guide the solution of increasingly difficult problems.  
  Color shows the model's drawing trajectory, from start (blue) to finish (pink). Panels \textbf{(D-E)} illustrate the most interesting dreams found across five runs, both before and after learning.
  Fig.~\ref{allLogoDreams} shows 150 random dreams at each stage.}\label{logoMaster}
  \end{figure}

  We next consider more creative problems: generating images, plans, and text. Procedural or generative visual concepts 
--- from Bongard problems~\cite{Moscow}, to handwritten characters~\cite{lake2015human,hofstadter1993letter}, to Raven's progressive
matrices~\cite{raven2003raven} --- are studied across AI and cognitive science, because they offer a bridge between low-level perception and high-level reasoning. Here we take
inspiration from LOGO Turtle graphics~\cite{turtle}, tasking our model
with drawing a corpus of 160 images (split 50/50 test/train; Fig.~\ref{logoMaster}A) while equipping it with control
over a `pen', along with imperative control flow, and arithmetic operations on angles and
distances.
After training \system for 20 wake/sleep cycles, we inspected the learned library (\ref{deepMontageAppendix}) and found interpretable parametric drawing
routines corresponding to the families of visual objects in its
training data, like polygons, circles, and spirals
(Fig.~\ref{logoMaster}B) -- without supervision the system %agent
has learned the basic types of objects in its visual world. It
additionally learns more abstract visual relationships, like
radial symmetry, which it models by abstracting out a new higher-order
function into its library (Fig.~\ref{logoMaster}C).

%What does DreamCoder dream of? 
Visualizing the system's dreams across its learning trajectory shows % provides a window into 
how the generative model bootstraps recognition model training: As the library grows and becomes more finely tuned to the domain, the neural net receives richer and more varied training data. At the beginning of learning, random programs written using the library
are simple and largely unstructured (Fig.~\ref{logoMaster}D), offering limited value for training the recognition model. 
After learning, the system's dreams are richly structured
(Fig.~\ref{logoMaster}E), compositionally recombining latent
building blocks and motifs acquired from the training data in creative %% (and sometimes crazy)
ways never seen in its waking experience, but ideal for training a broadly generalizable recognition model~\cite{tobin2017domain}.

\begin{figure}
  \centering\includegraphics[width = \textwidth]{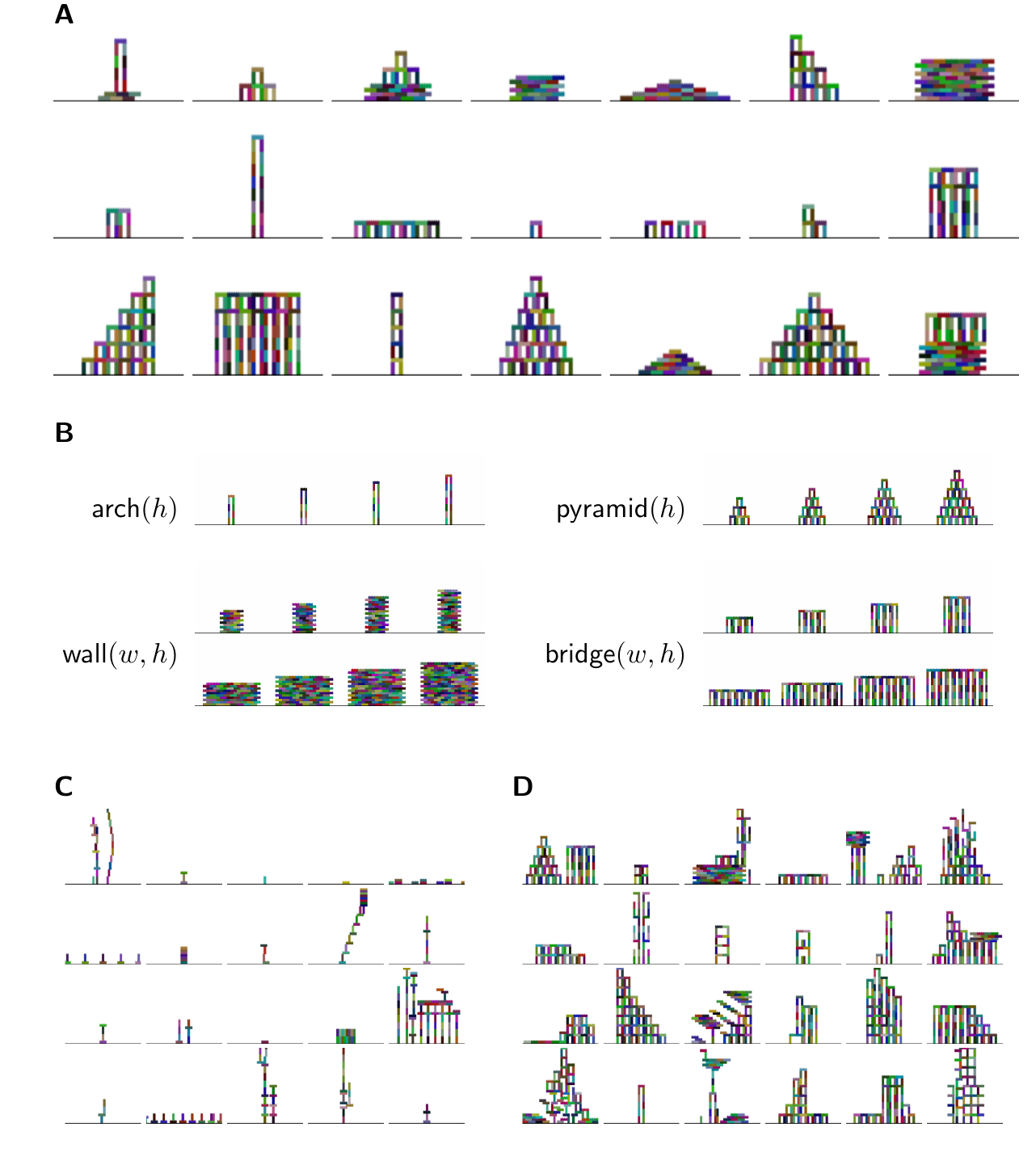}
  \dcaption{\textbf{(A)} 21 (out of 107) tower building tasks. The model writes a program controlling a `hand' that builds the target tower. \textbf{(B)} Four learned library routines. These components act like parametric options~\cite{sutton1999between},
    giving human-understandable, higher-level building blocks that the system can use to plan. Dreams both before and after learning \textbf{(C-D)} show representative plans the system can imagine building. After 20 wake-sleep iterations (\textbf{D}) the model fantasizes complex structures it has not seen during waking, but that combine building motifs abstracted from solved tasks in order to provide training data for a robust neural recognition model.
    Dreams are selected from five different runs; Fig.~\ref{allTowerDreams} shows 150 random dreams at each stage. %% After learning, the dreams recombine learned library concepts in new ways never seen in the training tasks, producing an infinite stream of data for the neural recognition model.
  }\label{tower}
\end{figure}

%% We next turn to a procedural planning domain, 
Inspired by the classic AI `copy demo' -- where an agent looks at
a tower made of toy blocks then re-creates it~\cite{towerCopy} -- we next gave \system 107 tower `copy tasks'
(split 50/50 test/train, Fig.~\ref{tower}A), where the system observes both an image of a
tower and the locations of each of its blocks, and must write a
program that plans how a simulated hand would build the tower.
%% These towers are built from Lego-style blocks
%% that snap together on a discrete grid.
The system starts with the same control flow primitives as with
LOGO graphics.
Inside its learned library we find parametric `options'~\cite{sutton1999between} for building blocks towers (Fig.~\ref{tower}B),
including concepts like arches, staircases, and bridges, %and brick walls
which one also sees in the model's dreams~(Fig.~\ref{tower}C-D).

Next we consider few-shot learning of probabilistic generative concepts, an ability that comes naturally to
humans, from learning new rules in natural language~\cite{marcus1999rule},
to learning routines for symbols and signs
\cite{lake2015human},
to learning new motor routines for producing words~\cite{lake2014one}.
We first %% investigate few-shot generative modeling by 
task \system with inferring a probabilistic regular expression (or Regex, see Fig.~1A for examples) from a small number of strings, where these strings are drawn from 256 CSV columns crawled from the web (data from~\cite{mws}, tasks split 50/50 test/train, 5 example strings per concept). The system learns to learn
regular expressions that describe the structure of  typically occurring text concepts,
such as phone numbers, dates, times, or monetary amounts (Fig.~\ref{qualitativeRegex}).
It can explain many real-world text patterns
and use its explanations as a probabilistic generative model to imagine new examples of these concepts. % (Appendix Fig.~\ref{qualitativeRegex}).
For instance, though \system knows nothing about dollar amounts it can infer an abstract pattern behind the examples \code{\$5.70}, \code{\$2.80}, \code{\$7.60}, $\dots$, to generate \code{\$2.40} and \code{\$3.30} as other examples of the same concept.
Given patterns with exceptions,
such as
%\code{q0005\_0003}, \code{q0009\_0003}, \code{q0009\_0008}, $\dots$, \code{q0002},
%train ['c04p0100a', 'c04p01007', 'c04p01009', 'c04p0100c', 'c04p01002']
%samples from generative model ['c04p0100c', 'c04p0100a', 'c04p01007', 'c04p0100c', 'c04p0100c']
\code{-4.26}, \code{-1.69}, \code{-1.622}, $\dots$, \code{-1}
it infers a probabilistic model that typically generates strings such as
\code{-9.9} and occasionally generates strings such as \code{-2}.
It can also learn more esoteric concepts, which humans may find unfamiliar but can still readily learn and generalize from a few examples: Given examples \code{-00:16:05.9}, \code{-00:19:52.9}, \code{-00:33:24.7}, $\dots$, it infers a generative concept that produces \code{-00:93:53.2}, as well as plausible near misses such as \code{-00:23=43.3}.

We last consider inferring real-valued parametric equations generating smooth trajectories (see~\ref{regressionSection} and Fig.~1A, `Symbolic Regression').
Each task is to fit data generated by a specific curve -- % family of curves --
either a rational function or a polynomial of up to degree 4.
We initialize DreamCoder with addition, multiplication, division,
and, critically,
arbitrary real-valued parameters,
which we optimize over via inner-loop gradient descent.
We model each parametric program as probabilistically generating a
family of curves,
and penalize use of these continuous parameters via the
Bayesian Information Criterion (BIC)~\cite{Bishop:2006:PRM:1162264}.
Our Bayesian machinery learns to home in on programs generating
curves that explain the data while parsimoniously avoiding extraneous continuous parameters.
For example, given real-valued data from $1.7x^2 - 2.1x + 1.8$ it infers a
program with three continuous parameters, but given
data from $\frac{2.3}{x - 2.8}$ it infers a program with two continuous parameters.
% shows example model outputs on held out generative modeling tasks.
%contrasting the full model with ablations that lesion either the library learning or recognition model training.

\subsubsection*{Quantitative analyses of \system across domains}\label{quantitative}
To better understand how \system learns,
we compared our full system on held out test problems with ablations missing either the neural recognition model (the ``dreaming'' sleep phase) or ability to form new library routines (the ``abstraction'' sleep phase).
We contrast with several baselines:
\emph{Exploration-Compression}~\cite{Dechter:2013:BLV:2540128.2540316},
which alternately searches for programs,
and then compresses out reused components into a learned library,
but without our refactoring algorithm;
\emph{Neural Program Synthesis},
which trains a RobustFill~\cite{devlin2017robustfill} model on
samples from the initial library; % for 24 hours;
and \emph{Enumeration},
which performs type-directed enumeration~\cite{feser2015synthesizing}
for 24 hours per task,
generating and testing up to
$400$ million programs for each task.
To isolate the role of compression %refactoring, 
in learning good libraries,
we also construct two \emph{Memorize} baselines.
These variants extend the library by incorporating task solutions wholesale as new primitives; they do not attempt to compress but simply memorize
%effectively memorizing
solutions found during waking for potential reuse on new problems (cf.~\cite{cropperplaygol}).
We evaluate memorize variants both with and without neural recognition models.

\begin{figure}[t]\centering
\vspace{-0.9cm}  \includegraphics[width = \textwidth]{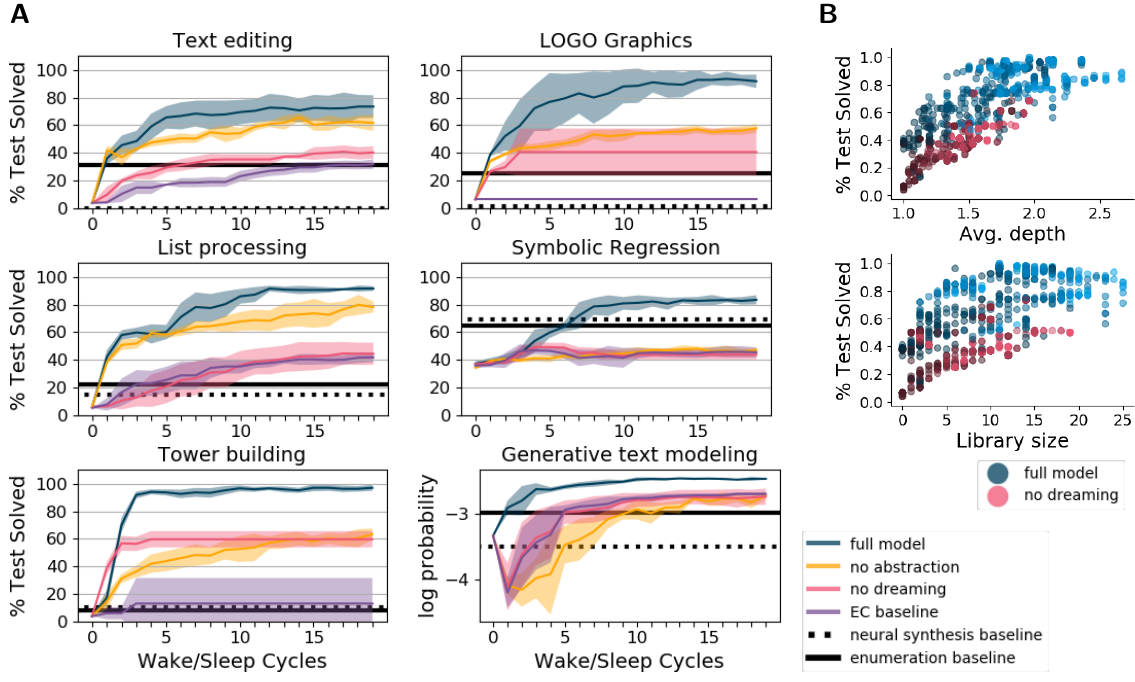}
\dcaption{Quantitative  comparisons of \system performance with ablations and baseline program induction methods; further baselines shown in Fig.~\ref{memorizeCurves}. \textbf{(A)}  Held-out test set accuracy, across 20 iterations of wake/sleep learning for six domains. %% Enumeration baseline: no learning but 24hrs of program enumeration per task. All other models given 10min timeout per task.
  Generative text modeling plots show posterior predictive likelihood of held-out strings on held out tasks, normalized per-character.
   Error bars: $\pm 1 \text{ std. dev.}$ over five runs.
    \textbf{(B)} Evolution of library structure over wake/sleep cycles (darker: earlier cycles; brighter: later cycles). Each dot is a single wake/sleep cycle for a single run on a single domain.
    Larger, deeper libraries are correlated with solving more tasks.
    The dreaming phase bootstraps these deeper, broader libraries,
    and also, for a fixed library structure, dreaming leads to higher performance.    
    %Test set performance across wake/sleep iterations. Each curve is a run with a different random seed. Teal: DreamCoder. Orange: Baseline, Neural Program Synthesis. Purple: Baseline, Exploration-Compression (EC).  Generative text modeling plots show posterior predictive likelihood of held-out strings on held out tasks, normalized per-character. %% \textbf{(B):} Deeper networks solve more testing tasks, both as measured by average and maximum depth (top/middle). Bigger networks also associated with solving tasks (bottom), but across domains the association is weaker. Full model (stars on scatterplots) learns bigger and deeper libraries than no-recognition-model ablation (crosses on plots). Each scatter plot point run with a different random seed.
  }\label{learningCurves}      
\end{figure}

Across domains, our model always solves the most held-out tasks (Fig.~\ref{learningCurves}A; see Fig.~\ref{memorizeCurves} for memorization baselines) and generally solves them in the least time (mean 54.1s; median 15.0s; Fig. S11).
These results establish that each of DreamCoder's core components -- library learning with refactoring and compression during the sleep-abstraction phase, and recognition model learning during the sleep-dreaming phase -- contributes substantively to its overall performance.
The synergy between these components is especially clear in the more creative, generative structure building domains, LOGO graphics and tower building, where no alternative model ever solves more than 60\% of held-out tasks while DreamCoder learns to solve nearly 100\% of them. 
The time needed to train DreamCoder to the points of convergence shown in Fig.~\ref{learningCurves}A % -- getting to the point where these problems can be solved in minutes or seconds --  
varies across domains, but typically takes around a day using moderate compute resources (20-100 CPUs). 
%%  -- with both components
%% combined, suggesting a synergy between these forms of declarative and
%% procedural learning.  

Examining how the learned libraries grow over time, both with and without learned recognition models, reveals functionally significant differences in their depths and sizes. Across domains, deeper libraries correlate well with solving more tasks ($r=0.79$), %($p < 10^{-186}$), 
and the presence of a learned recognition model leads to better performance at all depths. The recognition model also leads to deeper libraries by the end of learning, with correspondingly higher asymptotic performance levels (Fig.~\ref{learningCurves}B, Fig.~\ref{depthMontage}).  Similar but weaker relationships hold between the size of the learned library and performance. 
%
%, and these deeper libraries correlate well
%with solving more tasks ($r=0.79$). %%   Within each domain, the
%% presence of a learned recognition model leads to larger libraries of learned explicit concepts, although the correlation between the number of learned subroutine concepts and the number of solved tasks is weaker and not statistically significant ($r=0.20$, $p >
%% 0.05$).  
Thus the recognition
model appears to bootstrap ``better'' libraries, where ``better'' correlates
with both the depth and breadth of the learned
%% procedural learning.
%% Kevin December 23 version
%% We measure interactions between dreaming and the learned library's structure:
%% compared to the no-dreaming ablation,
%% the full model converges to deeper and broader libraries,
%% and these deeper libraries correlate well
%% with solving more tasks ($r=0.79$).
%% Simultaneously, for a given library depth/size,
%% dreaming leads to solving more tasks.
%% Thus, for a fixed library structure,
%% dreaming aids problem solving,
%% and also bootstraps ``better'' library structures,
%% where ``better'' correlates
%% with the depth and breadth of the learned
symbolic representation. % (see Fig.~\ref{learningCurves}B,Fig.~\ref{depthMontage}).

%NEW Version 1:
Insight into how DreamCoder's recognition model bootstraps the learned library comes from looking at how these  %implicit and explicit 
representations 
jointly embed the similarity structure of %the landscape of 
tasks to be solved. 
%Just as human experts learn to intuitively recognize avenues of solution even before solving a problem~\cite{chi1988nature},
%even \emph{before} solving a problem.
DreamCoder first encodes a task in the activations of its recognition network,
then rerepresents that task in terms of a symbolic program solving it. % guiding its search for solutions.
Over the course of learning, these implicit initial representations realign with the explicit structure of the final program solutions,
as measured by increasing correlations between the similarity of problems in the recognition network's activation space and the similarity of code components used to solve these problems %(RSA:~\cite{kriegeskorte2008representational}; 
(see Fig.~\ref{rsa}; $p < 10^{-4}$ using $\chi^2$ test pre/post learning).
Visualizing these learned task similarities (with t-SNE embeddings) suggests that, as the model gains a richer conceptual vocabulary,
its representations evolve to group together tasks sharing more abstract commonalities (Fig.~\ref{allTSNE}) -- possibly analogous to how human domain experts learn to classify problems by the underlying principles that govern their solution rather than superficial similarities~\cite{chi1981categorization,chi1988nature}.

\subsubsection*{From learning libraries to learning languages}\label{librariesToLanguages}

%basic -> library of more advanced routines to capture braod expertise. 
%
%could we also learn the basic building blocks?
%

%{\bf Version 1.} Our experiments up to now have studied how \system grows from a ``beginner'' state given only basic domain-specific procedures, such that only the easiest problems in a domain have simple, short solutions, to an ``expert'' state with a library of more advanced routines that allow even the hardest problems to be solved with short, meaningful programs.  Now we ask whether it is possible to learn starting from a more minimal state, absent even basic domain knowledge: Can \system start with only a generic basis of programming primitives, such that no problem in the domain initially has a simple solution, and grow a domain-specific language comprising both basic %primitives and advanced routines allowing the learner to solve all the problems in a domain?  We first consider learning a language for physical laws, starting with generic sequence-manipulation primitives typical of a modern functional programming language (such as Haskell), and then explore how such a programming language could itself be learned from a more minimal basis of recursive functions. %mention OCaml? 1950's LISP?

Our experiments up to now have studied how \system grows from a ``beginner'' state given basic domain-specific procedures, such that only the easiest problems have simple, short solutions, to an ``expert'' state with concepts allowing even the %domain's 
hardest problems to be solved with short, meaningful programs.  Now we ask whether it is possible to learn from a more minimal starting state, without even basic domain knowledge: Can \system start with only highly generic programming and arithmetic primitives, %% of programming primitives
%such that no problem in the domain initially has a simple solution, 
and grow a domain-specific language with both basic  
and advanced domain concepts allowing it to solve all the problems in a domain?  %We first consider learning a language for physical laws, starting with generic sequence-manipulation primitives typical of a modern functional programming language (such as Haskell), and then explore how such a programming language could itself be learned from an even more minimal basis. % of recursive functions. %mention OCaml? 1950's LISP?

% {\bf Version 2.} Our experiments up to now have initialized learning with a basic set of domain-specific procedures and studied how \system builds a library of more advanced routines representing deep domain expertise. Now we ask, could \system learn even the most basic building blocks of domain knowledge?  That is, could the system start with only a generic basis of programming primitives, such that no problem in a domain initially has a simple, short solution, and grow a domain-specific language comprising both basic and advanced routines allowing it to solve all the problems in a domain with compact, meaningful programs?  
%
%%OLD
%Rather than start with a domain-specific basis and enrich it over successive wake/sleep cycles -- as these prior experiments have done -- could we instead start with a highly generic basis, and then learn %a library encoding the basic domain language? 
%
%{\bf Optional: here or in discussion?} Acquiring expertise in this sense comes closer to the (hard) computational problem of learning a domain theory, which remains a central outstanding challenge in both cognitive science and AI. 
%

Motivated by classic work on
inferring physical laws from experimental data~\cite{simon1981scientific,langley1987scientific,schmidt2009distilling}, 
we first task \system with learning equations describing 60 different physical laws and mathematical identities taken from
AP and MCAT physics ``cheat sheets'', based on numerical examples of data obeying each equation. 
%and textbook equation guides,
The full dataset includes %including inverse square laws, expressions for ballistic motion,
data generated from many well-known laws in mechanics and electromagnetism, which are naturally expressed using
concepts like vectors, forces, and ratios.
Rather than give DreamCoder these mathematical abstractions, we initialize the system with a much more generic basis --- just a small number of recursive sequence manipulation primitives like
\code{map} and \code{fold}, and arithmetic --- and test whether it can learn an appropriate mathematical language of physics.
Indeed, after 8 wake/sleep cycles \system learns 93\% of the laws and identities in the dataset, %building a multi-level language of mathematical concepts to express their solutions (Fig.~\ref{librariesToLanguages}A). 
by first learning the building blocks of vector algebra,
such as inner products, vector sums, and norms (Fig.~\ref{librariesToLanguages}A).
It then uses this mathematical vocabulary to construct concepts underlying multiple physical laws, such as the inverse square law schema that enables it to learn Newton's law of gravitation and Coulomb's law of electrostatic force, %or the quadratic form underlying equations of ballistic motion and angular motion over time.  
effectively undergoing a `change of basis' from the initial recursive sequence processing language to a physics-style basis. % from the initial recursive sequence processing basis.

\begin{figure}
  \centering
% \vspace{-2.5cm}
 %\includegraphics[width = \textwidth]{figures/McCarthyPhysics_Combined_9_11.png}
 %\hspace*{-1cm}\includegraphics[width = 1.1\textwidth]{figures/figureSevenFinal.png}

\hspace*{-1cm}\includegraphics[width = 1.1\textwidth]{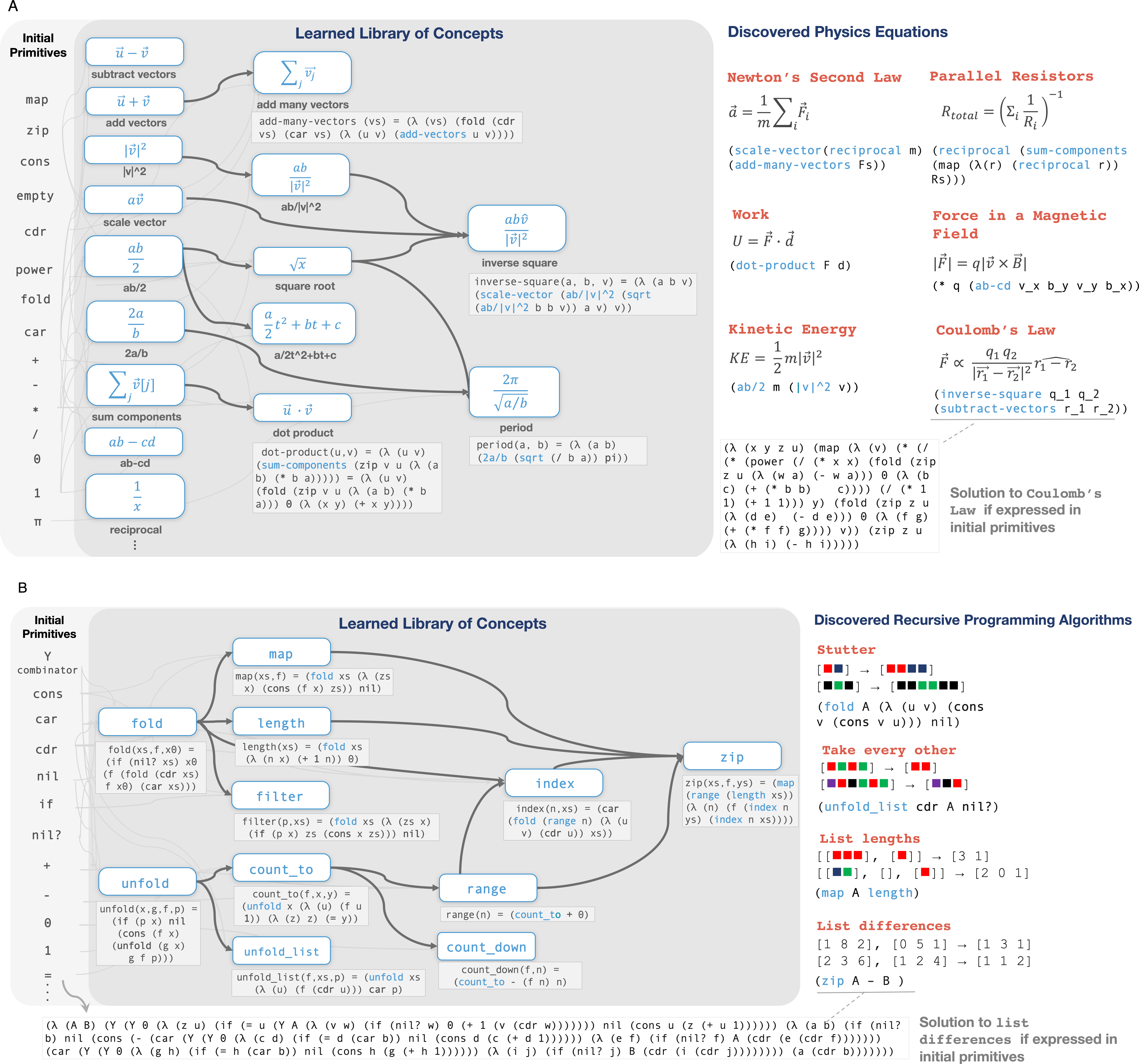}

  \caption{DreamCoder develops  languages for physical laws (starting from recursive functions) and recursion patterns (starting from the Y-combinator, \code{cons}, \code{if}, etc.) \textbf{(A)} Learning a language for physical laws starting with recursive list routines such as \code{map} and \code{fold}. \system observes numerical data from 60 physical laws and relations, %% , 20 input/outputs per task
     and learns concepts from vector algebra (e.g., dot products) and classical physics (e.g., inverse-square laws). Vectors are represented as lists of numbers. Physical constants are expressed in Planck units.  \textbf{(B)} Learning a language for recursive list routines starting with only recursion and primitives found in 1959 Lisp. DreamCoder rediscovers the ``origami'' basis of functional programming, learning \code{fold} and \code{unfold} at the root, with other basic primitives as variations on one of those two families (e.g., \code{map} and \code{filter} in the \code{fold} family), and more advanced primitives (e.g., \code{index}) that bring together the \code{fold} and \code{unfold} families.}%\vspace{-5cm}
  %For full tasks see Appendix~\ref{physicsAppendix} and~\ref{appendixMcCarthy}}.
  \label{librariesToLanguages}
\end{figure}

Could DreamCoder also learn this recursive sequence manipulation language? %which we provided as the initial basis for learning physics 
%not only allowed it to learn %vector algebra and 
%physics 
% 
%which formed the initial basis for learning vector algebra and more generally the foundation of modern functional programming?
%
%% This goal is relevant to a long-standing dream within the program induction community to ``learn from scratch'': starting with a \emph{minimal} Turing-complete programming language,
%% and then learning to solve a wide swath of
%% induction problems~\cite{solomonoff1964formal,schmidhuber2004optimal,hutter2004universal,solomonoff1989system}.
%% All existing systems,
%% including ours,
%% fall far short of this dream,
%% and it is unclear (and we believe unlikely)
%% that this dream could ever be fully realized.
%% How far can we push in this direction?
%% ``Learning from scratch'' is subjective, but a reasonable
%% starting point is a minimal subset of the primitives provided in 1959
%% Lisp~\cite{mccarthy1960recursive}: these include
%% conditionals, recursion, arithmetic, and the 
%% list operators \code{cons}, \code{car}, \code{cdr}, and \code{nil}.
%% A  basic first goal is to start with
%% these primitives,
%% and then recover a representation that
%% more closely resembles modern functional languages like Haskell and OCaml.
%
%Finally we consider learning the fundamental recursive sequence manipulation primitives,
%
%To investigate, 
We initialized the system %To answer this question we gave \system
with a minimal subset of
1959 Lisp primitives (\code{car}, \code{cdr}, \code{cons}, $\dots$) and asked it to solve 20 basic programming tasks, like those used in introductory computer science classes. %A key difference between this setup and our previous experiments is that %% , for this experiment,
Crucially the initial language includes
primitive recursion (the Y combinator), which in principle allows learning to express any recursive function, but no other recursive function is given to start;
previously we had sequestered recursion within higher-order functions (\code{map}, \code{fold}, $\dots$) given to the learner as primitives.
With enough compute time (roughly five days on 64 CPUs), \system learns to solve all 20 problems, % over several wake-sleep cycles, 
and in so doing assembles a library equivalent to the modern repertoire of functional programming idioms, including \code{map}, \code{fold}, \code{zip},
\code{length}, and arithmetic operations such as building
lists of natural numbers between an interval (see Fig.~\ref{librariesToLanguages}B).
All these library functions are expressible in terms of the higher-order function \code{fold} and its dual \code{unfold}, which, in a precise formal manner, are the two most elemental operations over recursive data --
a discovery termed ``origami programming''~\cite{gibbons2003origami}.
\system retraced the discovery of origami programming:
first reinventing fold,
then unfold,
and then defining all other recursive functions in terms of folding and unfolding.  %The end point of learning in this domain -- the recursive programming library shown in Fig.~\ref{librariesToLanguages}B -- essentially provides the starting point from which DreamCoder begins in learning vector algebra and physics in the previous case study, as well as the starting state for other domains in this paper, when supplemented by a small number of necessary domain-specific concepts in each case.  

\section*{Discussion}
Our work shows that it is possible and practical to build a single general-purpose program induction system that learns the expertise needed to represent and solve new learning tasks in many qualitatively different domains, and that improves its expertise with experience. 
Optimal expertise in \system hinges on learning explicit declarative knowledge together with the implicit procedural skill to use it. More generally, DreamCoder's ability to learn deep explicit representations of a domain's conceptual structure shows the power of combining symbolic, probabilistic and neural learning approaches: Hierarchical representation learning algorithms can create knowledge understandable to humans, in contrast to conventional deep learning with
neural networks, yielding symbolic representations of expertise that flexibly adapt and grow with experience, in contrast to traditional AI expert systems. 

We focused here on problems where the solution space is well captured by crisp symbolic forms, even in domains that admit other complexities such as pixel image inputs, or exceptions and irregularities in
generative text patterns, or continuous parameters in our symbolic regression examples. Nonetheless, much real-world data is far messier. A key challenge for program induction going forward is to handle more pervasive noise and uncertainty, by leaning more heavily on probabilistic and neural AI approaches~\cite{ellis2017learning,lake2015human,graves2016hybrid}.   Recent research has explored program induction with various hybrid neuro-symbolic representations~\cite{valkov2018houdini,andreas2016neural,manhaeve2018deepproblog,young2019learning,feinman2020generating}, and integrating these approaches with the library learning and bootstrapping capacities of DreamCoder could be especially valuable going forward.  

Scaling up program induction to the full AI landscape --- to commonsense reasoning, natural language understanding, or causal inference, for instance --- will demand much more innovation but holds great promise. As a substrate for learning, programs uniquely combine universal expressiveness, data-efficient generalization, and the potential for interpretable, compositional reuse.  Now that we can start to learn not just individual programs, but whole %libraries and languages for programming,
domain-specific languages for programming, a further property takes on heightened importance: Programs represent knowledge in a way that is mutually understandable by both humans and machines. Recognizing that every AI system %today 
is in reality the joint product of human and machine intelligence, % -- with humans invariably contributing what feels like the ``real intelligence’’ -- 
we see the toolkit presented here as helping to lay the foundation for a scaling path to AI that people and machines can truly build together.

In the rest of this discussion, we  consider the broader implications of our work for building better models of human learning, and more human-like forms of machine learning.

\subsection*{Interfaces with biological learning}

DreamCoder's wake-sleep mechanics draw inspiration from the Helmholtz machine, which is itself loosely inspired by human learning during sleep.
DreamCoder adds the notion of a pair of interleaved sleep cycles, 
and intriguingly, biological sleep similarly comes in multiple stages.
Fast-wave REM sleep, or dream sleep, is associated with learning processes that give rise to implicit procedural skill~\cite{fosse2003dreaming},
and engages both episodic replay and dreaming,
analogous to our model's dream sleep phase.
Slow-wave sleep is associated with the formation and consolidation of new declarative abstractions~\cite{DUDAI201520},
roughly mapping to our model's abstraction sleep phase.
While neither DreamCoder nor the Helmholtz machine are intended as  biological models,
we speculate that our approach could bring 
wake-sleep learning algorithms closer to the actual learning processes
that occur during human sleep.

DreamCoder's knowledge grows gradually, with dynamics 
related to but different from earlier developmental proposals for
``curriculum learning''~\cite{bengio2009curriculum} and ``starting small''~\cite{elman1993learning}.
Instead of solving increasingly difficult tasks ordered %deliberately 
by a human teacher (the ``curriculum''), 
DreamCoder learns in a way that is arguably more like natural unsupervised exploration: It attempts to solve random samples of tasks, searching out to the boundary of its abilities during waking,
and then pushing that boundary outward during its sleep cycles,
bootstrapping solutions to harder tasks from concepts learned 
with easier ones.
But humans learn in much more active ways: They can choose which tasks to solve,
and even generate their own tasks, 
either as stepping stones towards harder unsolved problems or motivated by considerations like curiosity and aesthetics.
Building agents that generate their own problems in these human-like ways is an important next step.

%\textbf{moving expertise to discussion? JOSH: I've tried moving a lighter weight version back to intro.}
Our division of domain expertise into explicit declarative knowledge and implicit procedural skill is loosely inspired by dual-process models in cognitive science~\cite{evans1984heuristic,kahneman2011thinking} and the study of human expertise~\cite{chi1981categorization,chi1988nature}. Human experts learn both declarative domain concepts that they can talk about in words -- artists learn arcs, symmetries, and perspectives; physicists learn inner products, vector fields, and inverse square laws -- as well procedural (and implicit) skill in deploying those concepts quickly to solve new problems.
Together, these two kinds of knowledge let experts more faithfully classify problems based on the ``deep structure'' of their solutions~\cite{chi1981categorization,chi1988nature}, and intuit which concepts are likely to be useful in solving a task even before they start searching for a solution.
We believe both kinds of expertise are necessary ingredients in learning systems, both biological and artificial, and see neural and symbolic approaches playing complementary roles here.

\subsection*{What to build in, and how to learn the rest}

The goal of learning like a human---in particular, a human child---is often equated with the goal of learning ``from scratch'', by researchers who presume, following Turing \cite{turing50}, that children start off close to a blank slate: ``something like a notebook as one buys it from the stationers. Rather little mechanism and lots of blank sheets.''  The roots of program induction as an approach to general AI also lie in this vision, motivated by early results showing that in principle, from only a minimal Turing-complete language, it is possible to induce programs that solve any problem with a computable answer~\cite{solomonoff1964formal,solomonoff1989system, schmidhuber2004optimal,hutter2004universal}. 
DreamCoder's ability to start from minimal bases and discover the vocabularies of functional programming, vector algebra, and physics could be seen as another step towards that goal. Could this approach be extended to learn not just one domain at a time, but to simultaneously develop expertise across many different classes of problems, starting from only a %generic 
single minimal basis? 
%programming substrate? %the most minimal basis? %autonomously carving out its own domains alongside its own domain-specific representations?  %We anticipate progress here by 
Progress could be enabled by metalearning a cross-domain library or ``language of thought''~\cite{fodor1975language,piantadosi2011learning}, %Growing 
%a single generic programmatic substrate for learning and reasoning, 
as humans have built collectively through biological and cultural evolution, which can then differentiate itself into
representations for unboundedly many new domains of problems.

While these avenues would be fascinating to explore, %we do not advocate such a blank-slate approach as the most efficient path to more general AI. 
trying to learn so much starting from so little is unlikely to be our best route to AI -- especially when we have the shoulders of so many giants to stand on.
Even if learning from scratch is possible in principle, such approaches suffer from a notorious thirst for data--as in neural networks--or, if not data, then massive compute: Just to construct `origami' functional programming, DreamCoder took approximately a year of total CPU time.   Instead, we draw inspiration from the sketching approach to program synthesis~\cite{solar2008program}. Sketching approaches consider single synthesis problems in isolation, and expect a human engineer to outline the skeleton of a solution. Analogously, here we built in what we know constitutes useful ingredients for %the more general problem of
learning to solve %many different
synthesis tasks in many different domains -- relatively spartan but generically powerful sets of control flow operators, higher-order functions, and types. We then used learning to grow specialized languages atop these foundations.  The future of learning in program synthesis may lie with systems initialized %not with the most minimal basis, but with rich 
with even richer yet broadly applicable resources, such as those embodied by simulation engines or by the standard libraries of modern programming languages.

This vision also shapes how we see program induction best contributing to the goal %future 
of building more human-like AI -- not in terms of blank-slate learning, but learning on top of rich systems of built-in knowledge. Prior to learning the domains 
%kinds of concepts
we consider here, human children begin life with ``core knowledge'': %in some form: genetically specified brain systems 
conceptual systems for representing and reasoning about objects, agents, space, and other commonsense notions~\cite{lake2017bbs, spelke1992origins, carey2011origin}.  We strongly endorse approaches to AI that aim to build human-understandable knowledge, beginning with the kinds of conceptual resources that humans do. %, whether acquired via manual engineering or metalearning methods.
This may be our best route to growing artificial intelligence that lives in a human world, alongside and synergistically with human intelligence.

\bibliography{main}
\bibliographystyle{unsrt}

\noindent\textbf{Supplement.} Supplementary materials available at \url{https://web.mit.edu/ellisk/www/dreamcodersupplement.pdf}. \textbf{Acknowledgments.}
We thank L. Schulz, J. Andreas, T. Kulkarni, M. Kleiman-Weiner, J. M. Tenenbaum, M. Bernstein,, and E. Spelke for comments and suggestions that greatly improved the manuscript. Supported by grants from the Air Force Office of Scientific Research, the Army Research Office, the National Science Foundation-funded Center for Brains, Minds, and Machines, the MIT-IBM Watson AI Lab, Google, Microsoft and Amazon, and NSF graduate fellowships to K. Ellis and M. Nye. %% All code and data will be made available on a public github site upon publication. {\bf Author contributions.}
%% K.E., J.B.T., and A.S-L. conceived of the model.
%% K.E., C.W., A.S-L., and  J.B.T. created the algorithm.
%% K.E., M.N., M.S-M., and C.W. ran experiments.
%% K.E., L.C., C.W., M.N., L.M., and M.S-M. implemented the software.
%% C.W., K.E., and J.B.T. designed illustrations.
%% L.H., K.E., M.S-M., and L.M. collected task data sets.
%% J.B.T. and A.S-L. advised the project.
%% K.E., J.B.T., and C.W. wrote the paper.
%% {\bf Competing interests:} The authors declare no competing interests. 

%\input{appendix.tex}
\end{document}